# New Ideas for Brain Modelling 3

Kieran Greer, Distributed Computing Systems, Belfast, UK.
http://distributedcomputingsystems.co.uk
Version 1.7

*Abstract:* This paper considers a process for the creation and subsequent firing of sequences of neuronal patterns, as might be found in the human brain. The scale is one of larger patterns emerging from an ensemble mass, possibly through some type of energy equation and a reduction procedure. The links between the patterns can be formed naturally, as a residual effect of the pattern creation itself. This paper follows-on closely from the earlier research, including two earlier papers in the series and uses the ideas of entropy and cohesion. With a small addition, it is possible to show how the inter-pattern links can be determined. A compact Grid form of an earlier Counting Mechanism is also demonstrated and may be a new clustering technique. It is possible to explain how a very basic repeating structure can form the arbitrary patterns and activation sequences between them, and a key question of how nodes synchronise may even be answerable.

## 1    Introduction

This paper considers a process for the creation and subsequent firing of sequences of neuronal patterns, as might be found in the human brain. The scale is one of larger patterns emerging from an ensemble mass, possibly through some type of energy equation such as entropy and a reduction procedure. The links between the patterns can be formed naturally, as a residual effect of the pattern creation itself. If the process is valid, then the pattern creation can be relatively simplistic and automatic, where the neuron does not have to do anything particularly intelligent. The pattern interfaces become slightly abstract without firm boundaries and exact structure is determined more by averages or ratios. If the process is based on an energy equation that reduces to a more stable state, it may require less intentional behaviour than, for example, creating clusters in the constructive way of deliberately linking between nodes. For a large-scale structure this might be preferable. This paper follows-on closely from the earlier research, including two earlier papers in the series and uses the ideas of entropy and cohesion. An example of the reactive-proactive comparison can be carried out using dynamic links [12] or an earlier Counting Mechanism [13]; but a new compact Grid form is then suggested that may be another general mechanism. The paper finishes with an implementation architecture, for the realisation and storage of knowledge and memory, as part of a general design, based on distributed neural components

The rest of the paper is organised as follows: Section 2 describes recent and new theory about the linking mechanisms. Section 3 compares the cognitive architecture with other research. Section 4 traces though some general classifiers, to show differences in how they would interpret data. Section 5 suggests a repeating process that can create and activate patterns as part of the architecture, including pattern resonance. Section 6 introduces another structure that uses neuron pairing to produce a more conscious type of signal. Finally, section 7 gives some conclusions and discussion on the work, including some implementation comparisons.

## 2 Linking Patterns

A key point to the theory is that a lot of the neurons inside a pattern link with each other. It is not the case that there is a single path from one neuron to the next, but there are lots of links between all of the neurons. The test theory of [2] also used this principle and it is also assumed in the tests of [9]. While not every node needs to connect with every other node; if most nodes connect with each other, it is quite a good way of defining the pattern shape. It could be imagined that the nodes still have connections to other places, but when excited, the pattern group will reinforce itself sufficiently to make it significant, while other more sparsely connected sets would possibly die out. It is a well-known phenomenon, as described in the related work section. It would also be the case that closer 'patterns' would prefer to connect with each other. This would then also refer to more similar concepts, when the whole process can be more consistent with state transitions and such things. The papers [9][7] have already considered the general mechanisms and used energy with entropy as the measures to describe how the patterns might form. This paper extends those equations slightly to include the inter-pattern connections.

## 2.1 Earlier Research

Earlier papers have also discussed pattern creation and linking, under different levels of complexity. The very first cognitive model [11] proposed a 3-level hierarchy of increasing complexity and functionality. The top level would cluster similar concepts (or neuron patterns) and then allow one group to trigger another group through a reinforcement link, created from time-based events. In Figure 1, for example, the top level of the original cognitive model is shown. A trigger probably means something firing in sequence instead of in parallel and represents a state change. This top level consists of higher level concepts, drawn as the octagonal shapes. They are higher level because they group together lower level patterns. The idea of a concept is abstract and so the term could ultimately reduce to a single neuron, which might not be very helpful.

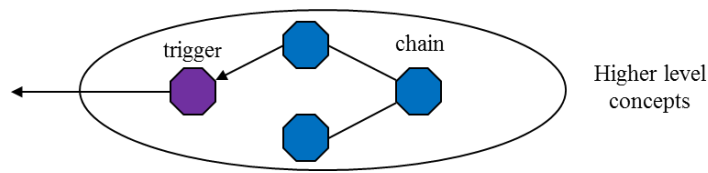

Figure 1. The Original Cognitive Model [11] top level clustering.

The model has been tested a certain amount, but it became clear that the trigger link between the clusters was not much different to a single node link. It would be created in essentially the same way, through reinforcement. As shown in Figure 2, this top level was expanded (Figure 4 of [9], for example) so that the other models of concept trees [8] and the symbolic neural network [12] could be incorporated. The concept trees can be considered as more knowledge-based, or even static memory structures. The symbolic neural network again represents symbols or concepts, but performs the same function as Figure 1, and would use time-based events as much as semantics, to dynamically link concepts. See the earlier papers for more details about this. These two structures have therefore been joined at a time-based events layer, where Figure 1 probably sits at that layer and above. This is indicated in Figure 2 with blue regions that circle the concepts that fire together. One can imagine a stimulus activating a number of these trees that then fire in parallel. This might then activate other patterns in sequence.

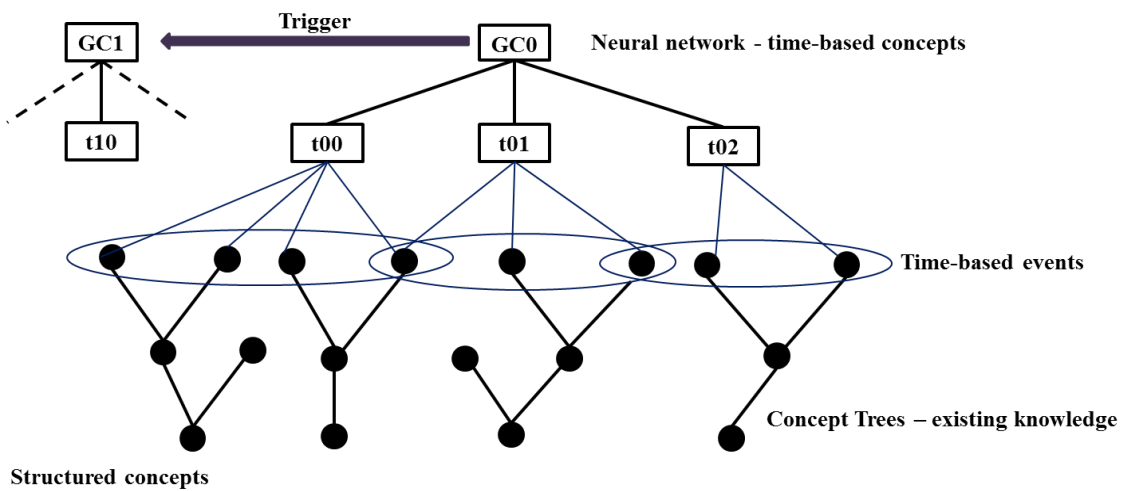

Figure 2. Concept Trees with the Neural Network Structure [9].

As the mechanisms are general, they can be repeated, depending on the scale for example, where a single concept is still a small pattern, and so on. A hierarchical structure is also assumed in most models [15], as is a difference between the physical space and the logical space. A hierarchy is useful because it can be a more economic structure and it can also perform a different function at each level, but for this paper, it is mostly about the structure. So Figure 2 combines dynamic experience-based at the top with static knowledge-based at the bottom. The very top GC global nodes in the neural network represent the accumulated branches and can trigger other global nodes. For state changes or firing differences to be recognised, a time factor has to be included. Let a source stimulus activate groups of neurons that then fire in parallel, but a state change might require the activation of patterns not initially connected with the stimulus. This is then also a sequential process as the energy source would come from the activated patterns and not the original stimulus, and could change the stimulus result. In Figure 3, for example, there are two connections between the patterns, which would cause activation at a slower rate than inside each pattern. This figure is used again in following sections.

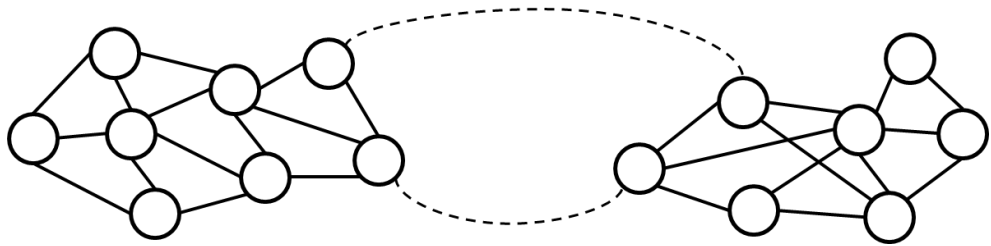

Figure 3. Two pattern ensembles with residual inter-pattern links between them.

## 2.2 Reactive Linking

Reactive linking may have more relevance where patterns sequentially and laterally activate each other, rather than try to form a hierarchy. Assume for example, that nodes in a region fire in some arbitrary manner. This would produce a larger mass or ensemble out of which distinct patterns can emerge and it is more of a reactive than a proactive process. These patterns are defined in physical space by the fact that their nodes have more links between each other than with other regions. A simplistic view of the reduction process is shown in Figure 3, where the patterns themselves are defined because they have more intra-pattern links and the inter-pattern links are the dotted lines between them.

If a pattern is mostly all connected, then one scenario is a small but persistent energy input that eventually stimulates all of the other nodes. Another scenario is a much larger stimulus that is able to trigger the pattern over a much shorter period of time. Therefore, depending on what patterns fire and the firing strength, this can cause other patterns to be more or less important, and be immediately realisable or realisable only through time.

## 3   Related Work

Physical systems are included in the related work, because this research is more interested in how global structures might change, not individual entities. Other topics are neural networks, bio-related and self-organising systems.

### 3.1   Global Properties

Weisbuch [30] describes the properties of a chaotic system inspired by statistical physics. One theory written about states that even if there is randomness or disorder at the local level, there may still be order at the global level. The disorder itself can repeat in a consistent way. The Physics model of percolation describes this phenomenon and allows for insulators or conductors between local nodes, but does not require a regular order at that level. He writes that:

'The simple models used by physicists are based on periodic networks, or grids, and simplified components of two different types are placed on the nodes, such as for example conductors or insulators in the problem known as percolation. These components are randomly distributed, and the interactions are limited to pairs of neighboring nodes. For large enough networks, we perceive that certain interesting properties do not depend on the particular sample created by a random selection, but of the parameters of this selection. In the case of the aforementioned insulator/conductor mixture, the conductivity between the two edges of the sample depends only on the ratio of the number of conductive sites to the number of insulating sites'.

Therefore, the interpretation of the result is as much about ratios or percentages than fixed boundaries. Each single interaction becomes only a small percentage of the desired result and therefore, if 'more' of the interactions are correct than incorrect at a global level, the desired result can be achieved. It may also be about the type of neuron behaviour. The idea of 'neurons that fire together wire together' is the well-known doctrine of Hebb (mentioned in [16]) and is central to how the patterns might form. Weisbuch notes that using Hebb's rule in a neural network will result in the network attractors representing the patterns. The system reduces to the state of the attractors, which are the sequences of repeating states. Memories can also be created this way.

Even if this research considers structure and not function, in biology, Interneurons are created for just that purpose [17] – for statistical connectivity and not functionality. Their results indicate that statistical connectivity can account for much of the specific synaptic patterning between neurons. Chemospecific steering is still essential and can tweak things, but it is non-specific – not aimed at any particular pair of neurons. They show that the exact positioning of neurons within their layer is not critical but that some intentional direction is possible. They note that the model would predict head-on collisions between neurons, where a chemspecific signal can steer axons and dendrites around them. The paper [23] is more computer-based and describes tests that show how varying the refractory (neuron dynamics) time with relation to link time delays (signal), can vary the transition states. They note that it is required to only change the properties of a small number of driver nodes, which have more input connections than others and these nodes can control synchronisation locally.

## 3.2 Earlier Research

Earlier research ideas that are relevant include [6] -[13] and are mostly concerned with how the system as a whole would work. The ReN [10] stands for Refined Neuron. It proposes an idea for re-balancing the network, by converting excess energy into new intermediary neurons. A signal from a source must activate the intermediary neurons first, before they activate the 'main' neuron, probably like the interneurons of [17]. This can refine the neuron signal values, because each source signal becomes fractional (not a whole 1 or 0) when it passes through the intermediate neurons first. With the ReN however, it is a proactive process, where surplus energy forces new connections to re-balance the system. Neurons firing together include the idea of resonance among the activated synapses (axons/dendrites), which then induces new paths to grow. The new neuron therefore represents some type of (sub)concept created by the firing group. And the process is necessary to re-balance the system and reduce entropy, or improve efficiency. If looking at the definition of entropy as part of thermodynamics, it is stated in terms of how much energy in a system is no longer available for conversion into mechanical work.

While pro-active mechanics can be localised, the theory of this paper also considers more reactive processes over a larger domain. The residual effect of a persistent general stimulus would keep certain patterns alive, but these can form in centres of attraction. A section of [7] is quite interesting and considers how the structure might be determined: 'It is easy to understand tree structures getting automatically narrower, but to broaden out requires the deliberate addition of new nodes and links to them. Re-balancing is always an option, where excess signal might encourage new neurons to grow, as in a ReN. Or many neuron clusters can interact and link with each other, but still provide specific paths into their own individual set of nodes. An idea of nested patterns might also help. Smaller or less important patterns at the periphery can be linked to by a more common mass in the centre, for example, leading to a kind of tree structure. In which case, it can be less of a deliberate act and more the residual result of a region being stimulated in a particular way.'

### 3.3 Biology-Related

Hawkins and Blakeslee [15] chapter 6, describe the brain cortex region as having 6 layers. A visual process, for example, uses 4 layers, where 3 layers form a hierarchy with the fourth (IT) layer joining this up to complete a circuit and recognise the pattern set as a significant. The circuit completion also helps the signal to say active. The author's current model is at a slightly coarser scale than this, but it is possible to see this type of hierarchy in it. As shown in Figure 2, the concept trees may be the indexes to a shallow Concept Base [12] hierarchy, represented by the blue time-based layer. That layer could also compare with the top hierarchy layer in the visual cortex, for example. The closing (IT) nodes of the visual cortex would then be the first level of the symbolic neural network, as is shown in the figure. This is of course, not intentional, but at least the architectures can show some resemblance to each other. The columnar nature of the firing that is explained is also interesting and mentioned again later.

The work of Tkačik et.al. [26] could be relevant. It uses statistical mechanics to try to explain some of the mechanisms that occur in the biological brain. They note that while the brain is not a thermodynamic system, it exhibits some of those properties and so it can be measured that way. They also note that the brain is a nonequilibrium system and ask the question of how it then obtains equilibrium (ReN). If dealing with a critical state $E$, they suggest that you can count the number of potential states that have energy close to or less than $E$ and consider those as the degrees of freedom or entropy. They then show how their results suggest a thermodynamic limit to the neural activity, but have no definite explanation of why. The limit suggests a boundary, but a more closed system can be maintained if activity is mostly inwards and helped if inhibitors also switch off the surrounding area. Their results also show that energy and entropy are linear with respect to each other. The equation 4 of that paper measures the entropy and it is similar in structure to the one used in this model ([9] section 4.1, or [7] equ. 9) to measure the activity of mostly inter-connected neural masses. They measured spiking events of individual neurons and so particular structures were not described that can be compared with, but potential states with slightly less energy might be like the next level in a tree. Another paper [24] further quantifies the pairwise correlations over larger networks and this whole biology-related section is important for a new structure, written about in section 6.

Also interesting are the theories of Neural Oscillations and Neural Binding[1]. These brain theories are used to help to explain how the brain is able to comprehend, what it sees, for example and therefore overlaps with thought and consciousness. With one visual system theory (Gray, Singer and others [25]), synchronous oscillations in neuronal ensembles bind neurons representing different features of an object. Gestalt psychology is also used, where objects are seen independently of their separate pieces. They have an 'other' interpretation of the sub-features and not just a summed whole of them. Chen uses the idea that perception depends on 'topological invariants that describe the geometrical potentiality of the entire stimulus configuration'. The idea here is that looking at a certain object creates a stimulus which in turn fires a constellation of that object, and then the object becomes

---

[1] See Wikipedia, for example.

recognized. The rebuttal is the fact that independent constellations would require too many neurons and pathways [20]. Singer and Gray write that:

'Representation of a feature by a population of cells raises binding problems when nearby contours evoke graded responses in overlapping groups of neurons. Of the many simultaneous responses, those evoked by the same contour need to be distinguished and evaluated together to avoid interference with the responses elicited by neighboring contours. A similar need for response selection and binding arises in the context of perceptual grouping.
…
Essentially, one cell would be required for every distinguishable feature, for each higher-order feature combination, and ultimately for every distinguishable perceptual object. Moreover, because of its inherent lack of flexibility, such a mechanism cannot easily cope with the representation of new or modified patterns.'

They also note that shared patterns would have many advantages, especially with neuron numbers, but a flexible process is required to clearly recognise when parts of a pattern fire. They then go on to write:

Simulation studies by Softky & Koch (1993) suggest that the interval for effective summation of converging inputs is only a few milliseconds in cortical neurons. Thus, if synchronization of discharges can be achieved with a precision in the millisecond range, it can define relationships among neurons with very high precision. Moreover, if synchrony is established rapidly and maintained only over brief intervals, different assemblies can be organized in rapid temporal succession.

## 3.4   Neural Models

Neural networks are the most obvious type of distributed model. The paper [4] is interesting and may even be trying to implement something similar in hardware. They note that electrical synapses are bi-directional, unlike the chemical ones, but this means an intentional signal in both directions and not an excess one. The term small-world effect [29] has similarities with pattern groups connected by residual links and is described as part of their new Hopfield network [18][1] design:

> 'It is known that brain neural network is comprised of millions of neurons and their connections. Major connections are of short distance and the number of long-distance connections is less, which is acclaimed as the so-called small-world effect. Recently, the small-world neural network model was thus inspired and proposed.
> …
> More specifically, it has greater local interconnectivity than a random network, but the average path length between any pair of nodes is smaller than that of a regular network. The combination of large clustering and short path length makes it an attractive model capable of specialized processing in local neighborhoods and distributed processing over the entire network.'

The Hopfield neural network, and its stochastic equivalent are auto-associative or memory networks. With the memory networks, information is sent between the input and the output until a stable state is reached, when the information does not change. These are resonance networks, such as bidirectional associative memory (BAM), or others. But they can only provide a memory recall of the data that was input and are also constrained in size to that amount of data. If some of the input pattern is missing however, they can still provide an accurate recall of the whole pattern. They also prefer the data vectors to be orthogonal without overlap. After the different stable states have been learned, they represent a type of energy function. A new pattern may be associated with a set of the state vectors, but one state will be closer to it, which will produce the best match.

Hopfield networks are also attractive because the units can operate asynchronously of each other. Each unit can compute its excitation at random times and change its state independently of the others. Also described in [22], chapter 13:

> Hopfield's approach illustrates the way theoretical physicists like to think about ensembles of computing units. No synchronization is required, each unit behaving as a kind of elementary system in complex interaction with the rest of the ensemble. An energy function must be introduced to harness the theoretical complexities posed by such an approach.

But while the learning can be asynchronous, the weights and links are still very directed, although self-organisation and an energy function can maybe take precedence over a supervised learning approach. Watts and Strogatz [29] then took the bi-directional links and replaced it with a circular architecture to produce the small-world networks. More importantly, the nodes are not all linked with each other, or completely randomly linked, but somewhere in-between. The linking pattern would still provide more group support than a random setup, for example. Considering Figure 2 and Figure 3 of this paper, is it possible to see small-world networks at the time-based events layer? If not, then it is still another look-alike structure to relate to, as possibly is [23].

Other types of model include Cognitive Architectures. The paper [19] is very interesting and uses a more knowledge-based vernacular to describe other cognitive systems. For example, where the current research has used the terms knowledge or experience-based, that paper may use the terms declarative or procedural knowledge, to mean probably the same thing. The paper also describes that solutions to the problem have to implement prototypes, exemplars and/or theory-theories structures. It shows that a number of other cognitive systems have implemented these artefacts and that a general theory needs to accommodate them. The main problems that the paper notes are the knowledge homogeneity problem (what heuristic to choose) and the integration of common-sense contextual knowledge, which are both required to allow more generality in the model. It will be shown in section 7.1 that the author's model can also be described using the more knowledge-based sets of terms and that the current work can accommodate the required transformations, to allow the information to flow seamlessly, from pattern-based data into knowledge.

# 4   Cluster Interpretations – Global Patterns vs Unique Instances

This example shows how even very basic reinforcement mechanisms can produce different interpretations of pattern sequences. Three types of count updated were tested, where the first two used more local count reinforcements, described by the following equations, also used in [7]. The idea of a pattern is the whole input set here:

$R_{ipt+1} = R_{ipt} +- \omega_I$ ............... Equation 1

$CI_{ipt+1} = CI_{ipt} + \omega_I \quad \forall \ ((N_i \in IP_t) \land (N_i \in P_p))$ ............... Equation 2

$CG_{ipt+1} = CG_{ipt} + \omega_G \quad \forall \ (N_i \in P_p) \land ((N_j \in IP_t) \land (N_j \in P_p))$ ............... Equation 3

Where:

$R_{ipt}$ = reinforcement or weight value for node *i* in pattern *p*, at time *t*.
$CI_{ipt}$ = total individual count for node *i* in pattern *p*, at time *t*.
$CG_{ipt}$ = total group count for node *i* in pattern *p*, at time *t*.
$P_p$ = pattern P.
$IP_t$ = input pattern activated at time t.
$N_i$ = Node i.
$\omega_I$ = individual increment value.
$\omega_G$ = group increment value.
$N_g$ = total number of group updates or events.

Consider the following example: There is a set of nodes A, B, C, D, E, F and G. Input patterns activate the nodes, which is then presented to the classifier as follows:

1. A fires with B, C, D and E.
2. B fires with A, C and D.
3. C fires with A, B and D.
4. D fires with A, B and C.
5. E fires with A, F and G.
6. F fires with E and G.
7. G fires with E and F.

## 4.1   Single Variable Reinforcement

With basic reinforcement, a weight is incremented when a variable is present and decremented when it is not, and can use Equation 1 to do this. If considering each variable as independent without any containing structure and a decrement value of 0, then the first set of counts, shown in Table 1 would occur. The whole group A-G is updated as a single entity and this would suggest that the best cluster group is B, C, D and E, with another one probably F and G, but those are not the best clusters.

|   | A | B | C | D | E | F | G |
|---|---|---|---|---|---|---|---|
| I | 5 | 4 | 4 | 4 | 4 | 3 | 3 |

Table 1. Count reinforcement when updating for individual variables.

### 4.2 Counting Mechanism with Unique Instances

A counting mechanism [12][13] has been suggested previously, where a weight is incremented when a variable is present and a group-related weight is incremented when any variable in the group is present, as described by Equation 2 and Equation 3. There are therefore two counts: the global count (G) is updated for every pattern presentation and the local count (I) is updated only when the variable itself is presented. If each pattern instance is presented separately and updated separately however, then a different result is shown in Table 2. Using some simple rules that measure how close the global and local counts are to each other - it is possible to calculate that the correct groups are A, B, C, D and E, F, G. Comparing the two counts shows that the difference between them is the least for those two groups, suggesting more coherence in those two groups.

|   | A, B, C, D, E | A, B, C, D | E, F, G, A | E, F, G |
|---|---|---|---|---|
| I | 1 | 3 | 1 | 2 |
| G | 7 | 4 | 3 | 2 |

Table 2. Counting Mechanism for unique group instances.

So why is this a better result than the individual reinforcement value of Table 1? Firstly, as the data is categorical or symbolic, it is not a problem of reducing a numerical error, but of matching over the data rows. Secondly, the first method does not consider the structure in the input patterns. If all 7 events are thought to be a single event, then the nodes B, C, D and E occur the most often and the same number of times. But if the rows are considered to be distinct, then this particular tuple does not occur very often, only once and probably also with A. The structure indicates that E definitely belongs with F and G. Therefore, if the input structure is also considered, it becomes more obvious why Table 2 is better. Why is the global count useful?[2] It simply makes it easier to see when the distinct row occurred versus when only parts of it occurred and is therefore a measure of coherence. As with the residual links between neural patterns, there is still overlap between the two groups which is not very clear in Table 2. This is written about again in the conclusions. The next section shows a more global form of the localised counting mechanism, converting it into a grid format and allowing for the inter-links to be seen more clearly.

### 4.3  Grid-Based Frequency Counts

This section describes a new grid-based frequency count that is more global in nature. It is clear from the data that A, B, C and D all reinforce each other (pattern 1), as does E, F and G (pattern 2), but there is still an inter-pattern link between A and E (in both pattern 1 and 2). With a grid format, the input is represented by a single pattern group, but this time the counts for each individual variable and included and cross-referenced, allowing the inherent structure to be included, as shown in Table 3. This is in fact probably a statistical matching equation more than an entropy one. The grid format lists each variable both as a row and a column. Each time a pattern is presented, the related cell value for both the row and the column is incremented by 1. The grid result can then be read using the following algorithm:

1. Each row displays count values representing a key variable - the row name, and its relation to the other variables.
2. All cells relating to variables in the input pattern are updated each time.

---

[2] Only useful not essential, because a single reinforcement value can produce a similar result, depending on what is updated each time.

a. Each row that starts with one of the variables updates the count for every other variable in the input pattern.
    b. Because the variable is repeated in several cells, this still leads to normal count values for each cell.
3. To determine the best clusters then:
    a. For a key variable (row key value) scan across and select the other variables with the largest count values. That variable then considers those other variables to be part of its cluster.
    b. The other variables however may be more associated with a different cluster, so their rows can be checked for consistency. They should similarly have a largest count value for the other variables in the cluster. If any have different (larger or smaller) count values, then they probably belong to a different cluster.
4. These different counts can still be considered for linking between patterns, where they have some association but belong to a different pattern.

|   | A | B | C | D | E | F | G |
|---|---|---|---|---|---|---|---|
| A | x | 4 | 4 | 4 | 2 | 1 | 1 |
| B | 4 | x | 4 | 4 | 1 | 0 | 0 |
| C | 4 | 4 | x | 4 | 1 | 0 | 0 |
| D | 4 | 4 | 4 | x | 1 | 0 | 0 |
| E | 2 | 1 | 1 | 1 | x | 3 | 3 |
| F | 1 | 0 | 0 | 0 | 3 | x | 3 |
| G | 1 | 0 | 0 | 0 | 3 | 3 | x |

Table 3. Display of the reinforcement between pattern presentations grouping A, B, C and D, plus E, F and G, with a single inter-pattern link A-E.

With the example set of events, the nodes would also be beside each other in the grid, but the variable placings in the grid can be mixed-up and still produce the same result. In row A, for example, the counts suggest that it should be clustered with B, C and D, which is the same cluster conclusion for rows B, C and D. It is probably not necessary to update a self-reference in the grid, so the leading diagonal can be empty. This grid format is possibly a transposition of the counting mechanism. In the Counting Mechanism [13], the global count represents some global pattern, or the global count for a unique pattern bit, while with the grid, global structure is implicit through the cross-referenced counts. The counting mechanism updates instances separately, but in the model, it still tries to cluster those instances together. The grid starts with a full set of variables and separates them into similar counts, although it might be possible to add variables to the grid dynamically, which the counting mechanism also accommodates. As the grid does not rely on prior classifications of the categories, it is really a self-organising mechanism for categorical data.

### 4.4  Test Examples

One use for the grid structure could be collaborative filtering, such as the Entree Chicago Recommendation Data Set [3][5]. With that data, the action or request of a user was coded and interpreted as a category. Then each user session was a group of related categories. A test program was then able to count the frequencies of categories that occurred together, as in the grid example, and create clusters of categories that more commonly occurred together. These can even be related back to the restaurant or user, but there are no clear results to compare with. Another test that did produce results was the Plants dataset [27], where the success was measured by regional clustering. It was used in [14] to mine for association rules, which are similar in the sense that the data can be seen as a set of binary events to associate together. With the plants dataset, different species in each state of USA and Canada are listed and the states were clustered on similar species. The results, given in Appendix A, do appear to show clustering based on geography or climate. One problem was highlighted. If a category ended up by itself, not clustered with any other category, then with the current algorithm, it is not obvious how to cluster it with another category or group. A more detailed examination would be required to give a good suggestion for that. However, the clustering process is a one-pass process and took only a few seconds to complete, on a standard laptop.

### 4.4.1 Comparing Grid with Counting Mechanism

All 3 methods of this section were used to cluster the plants dataset. A first test considered each input pattern simply as a set of variables and updated the variable's count each time it was presented. This appeared to give the least useful result, where the clustering was the most fragmented and included North with South, or East with West states in the same cluster. A second test firstly ordered the input dataset into decreasing numbers of variables and considered updating only from the same nested group. This helped a bit, but the nesting meant that most input patterns were still in a single group. The outer-most group would be added first and all existing variables would have their counts updated each time. The clustering was quite good but still with some geographic fragmentation. For the second test the global count was essentially redundant, because the dataset is being treated essentially as a single entity. A third test used the grid format and performed the best with regards to regional clustering, as shown in Appendix A. Basic count reinforcement (counting mechanism for example) would still cluster, but as well as the fragmentation in the clusters, it is more difficult to use, because value bands for each cluster need to be decided on manually, whereas with the grid, it is automatic. So with the counting mechanism, the suggestion is to use it more locally and create a new instance for each unique pattern that is presented.

## 5 Repeating Structures and Sequences

It is now clear that a general process can be described for how patterns may form and link with each other. It uses a structure that can repeat at different levels of granularity and can form in an automatic and arbitrary way. It is also completely mechanical in nature, requiring no sense of real intelligence. Certain themes repeat throughout the research, giving some confidence to their relevance. At the finest level of granularity, there can be single neurons with links between them. These can group together to form a pattern and weaker or residual sets of links can be left between patterns. If considering Figure 1, for example, it would be possible to replace the octagonal nodes with a pattern of neurons and the single links with the residual set. That whole figure can also represent a higher-level concept that triggers other ones through other links. So why not make the triggers a residual set of links as well, but between larger concept groups.

The nesting and time element in the structure is also obvious. For example, if an area of the brain is activated, then statistically, one pattern may fill with signal before another pattern. After a while, the pattern can activate an associated one through the residual set of links, but as that is a weaker connection, it would occur after the pattern itself is activated. The pattern may also throw inhibitors out into its environment, to switch off neighbouring patterns and make it the dominant one. It may also contain nested structures [9][7] that would represent sub-concepts and as the whole region fires it activates some of the sub-regions. This is an inward activation sequence, so to be completely successful, longer lateral connections with other global regions are also required. Therefore, lateral links between pattern groups occur as well as nesting.

## 5.1  Synchronising Nodes

This section adds some information about the known biological phenomenon that neurons tend to be evenly spaced [21] and there is a very simple mathematical theory to support it in [9]. An even spacing makes a lot of sense when considering node synchronisation inside patterns. If a root node activates a number of linked nodes, then if they are at an equal distance from the root node, they will all fire at the same time. If they subsequently do the same, then the nodes that they link to will also fire at the same time. Therefore, it is not just the fact that closer nodes cluster together, but rather that nodes of an equal distance from each other cluster into patterns. It is also helpful if the linked nodes feed back into the pattern again, so that they can activate the other pattern nodes with the same time interval and the whole firing sequence will be synchronised. The feedback links would not only define the pattern shape but also determine the firing rates, which would help the pattern to resonate. With so many connections however, saturation would be likely and so that would need to be managed. Another set of nodes may also be evenly spaced, but at a different distance, when they would synchronise differently and fire at a different rate. If the distance between the nodes is compared with a wavelength or maybe a frequency, then with different pattern groups in the same region, the brain can receive variable signals and the combinations could help with ideas about thought or consciousness.

### 5.2 Hierarchy Construction

While unique pattern instances were used previously with the neural network [12], that would be inefficient in a real brain model and so it may be possible to link the instances up thereby creating a unique hierarchy instance. For example, if nodes A-B-C-D get presented then they form a pattern. If A-B-C-D-E gets presented then create a new E node and link it to the A-B-C-D pattern. If over time, the A-B-C-D-E pattern occurs more often, then as with the concept trees, make it the whole pattern and remove the link. If A-B-C gets presented more often, then as with other papers [7], split the pattern because it is not cohesive enough. If something like A-F-G gets presented, then maybe create a new unique instance with a new 'A' node. An energy equation could suggest that it is easier to create a new unique instance than try to merge with an existing pattern. Maybe only node 'A' of the current pattern is required and afterwards, the patterns tend to act as a single unit. This type of clustering is still consistent with Table 2, for example, where maybe pattern 5 would have produced a whole new set of nodes.

## 6 Neuron Binding

The author would like to introduce another structure here, shown in Figure 4, that is new to the model. It would be another refinement that would help with the idea of consciousness and the merging of distinct patterns into a more singular whole. Biology has already suggested theories about neural oscillations and binding that include neural pairing. The pairing helps to group the neurons into specific patterns that the brain can understand, when different features can become synchronised and oscillate together. This new structure also relies on pairing neurons and may be loosely related to the earlier biological research. With this, two neurons would be used to represent an entity. One neuron exists in a base ensemble that is a flat structure and the other would exist in a hierarchical structure. Neural duplication is well-known about, but this structure would have duplication at a very close range. To find this structure therefore, some of the base neurons would be activated and probably realise an ensemble activation. That gives a much larger area to stimulate initially, when the hierarchical structure then provides more meaning to the pattern signals.

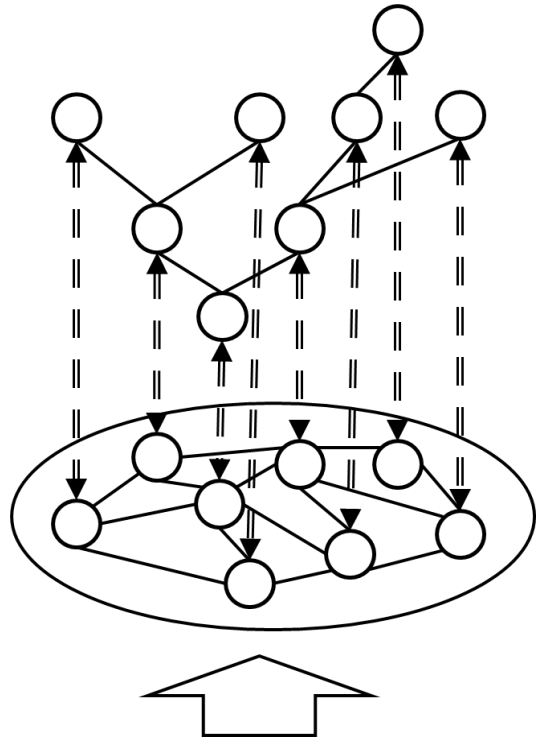

Figure 4. Neuron Pairing that can 'sing' through different resonance frequencies.

Therefore, the base ensemble fires and activates neurons. If these are paired with neurons in the hierarchical structure, then the tree structure is also activated. The hierarchy would send its signal on to somewhere else as well, such as the top-level neural network. The tree gives more definition and while it would be easy to see an ensemble mapping to a hierarchy in any brain model, the exact neuron mapping would be more difficult to explain. Some more details about this simple structure are given in [6]. As with general theory, the enclosing ensemble can re-activate parts of the neural tree. So, there can be a temporal and controlled element to the signalling process. The repetition of the mechanism through cycling, would also help to stabilise a changing signal pattern. So, to summarise: there are a number of local circuits that can be managed through the neuron pairs, as well as the larger enclosing patterns that can consistently reinforce and sustain a signal and the hierarchy provides some structural meaning, while the base ensemble makes the region easier to find.

## 6.1 Biology-Related

The two related work papers from the statistical mechanics background [24][26] are actually able to provide evidence to support this type of structure. A section from Tkačik et.al. [26] is as follows:

> … we expect the distribution of activity across neurons to reveal structures of biological significance. In the small patch of the retina that we consider, no two cells have truly identical input/output characteristics (44). Nonetheless, if we count how many combinations of spiking and silence have a given probability in groups of N >20 cells, this relationship is reproducible from group to group, and simplifies at larger N. This relationship between probability and numerosity of states is mathematically identical to the relationship between energy and entropy in statistical physics, and the simplification with increasing N suggests that we are seeing signs of a thermodynamic limit. If we can identify the thermodynamic limit, we can try to place the network in a phase diagram of possible networks. Critical surfaces that separate different phases often are associated with a balance between probability and numerosity: States that are a factor F times less probable also are a factor F times more numerous …

The larger area can therefore be defined more by probability and statistics, while the smaller area needs to consider exact structures. It is interesting that local spiking is distinct, but repeats on a larger scale. It is known that neurons are spaced evenly, but maybe small-scale differences are important, although there are also different cell types with different functionality. The idea that less probable events require larger numbers of them to balance the entropy is interesting, but is this simply a statistical requirement for the state transitions? A quote from Schneidman et.al. [24] is as follows:

> … We conclude that weak correlations among pairs of neurons coexist with strong correlations in the states of the population as a whole. One possible explanation is that there are specific multi–neuron correlations, whether driven by the stimulus or intrinsic to the network, which simply are not measured by looking at pairs of cells. Searching for such higher order effects presents many challenges (22–24). Another scenario is that small correlations among very many pairs could add up to a strong effect on the network as a whole. If correct, this would be an enormous simplification in our description of the network dynamics…

It is argued by both of these papers that the sparse nature on the neural firing means that pairwise correlations should occur much less often than is observed. If the correlation is hard-coded however, then it is easy for it to occur and the pairing can make related counts more linear in nature, although there is still the relation between all of the neuron pairs. The correlation is also an oscillating electrical signal. Another interesting comparison comes from [15] that describes a columnar architecture in the cortex, which means that either the tree structure is missing or there is an equally wide base from something else. The neuron pairing of Figure 4 however is only an idea for realising a type of phenomenon that is not easily explained. It would be even better if the spiking rates were in some part due to the distances between each neuron in the pair. The two papers go into a lot more detail and were not written with this structure in mind. Reading them should provide useful background information.

## 7 Conclusions and Discussion

This paper has described how patterns may form and more importantly link with each other through a general reduction process of a larger ensemble mass. The process can make use basic statistics and entropy equations [7] and here, the process can consider inter-pattern sequences. The signal strength inside of each pattern is stronger than between patterns, which helps to define it. However, time can be used to build up a signal, or more or fewer of the connections can be firing at the same time and so ratios or percentages are key. The influence of the input stimulus and the desired result are also factors. To show how the patterns and inter-pattern links might be realised, a Grid view of a global ensemble, updated by a series of pattern presentations has been demonstrated. The Grid format works well with event-based or categorical input and successfully clustered some plant data. The clustering is slightly different to the more traditional methods and so it may be a new general way of clustering, or specifically for this type of problem. A basic count reinforcement can also produce clusters in the plants dataset, but it was more fragmented when it came to the geographic locations. Considering the cognitive model of Figure 2: if the concept trees in the lower half represent rote-based learning in some way, they also represent distinct concepts. The time-based layer would be presented with groups of these and may re-cluster them into a different view that includes those time-based events. Then as with Gestalt theory, the experience-based neural network would receive the alternative view that may be more than just one (knowledge-based) concept at a time and it would change that view further, as it re-clustered to suit the experience-based needs. The grid format is making a frequency measurement over the whole dataset and not a similarity measure. It may therefore be an option for the time-based layer, to cluster everything into the second view. The counting mechanism and related methods are maybe better locally, with more clearly defined concepts.

The pattern-forming reduction process is much less organised than the concept trees or the symbolic neural network, but that is OK, as it defines a larger area that uses probability to overcome errors on a smaller scale. The trigger that is indicated in the diagrams is similar to a basic link, except that it is between two sets of patterns, not single nodes. As the structure gets smaller however, it becomes more precise and a hierarchy can be created by repeating the clustering at different levels. The unique instance of earlier papers might need to be replaced, for economic reasons, with instances being linked together and further merged or separated. How exactly this might influence the more dynamic, symbolic neural network at the top, is yet more research. Another process that occurs in biology is also useful. If a neuron is firing more often, then it is more likely to create a new neuron. Therefore, the terms with the largest values should be realised first in the concept trees as the base nodes. The next largest values should form the next level and so on, where it only remains to link up the different levels with branches. The root nodes are therefore also statistically the most important. So, this process is clearly seamless and has transformed a neural mass into a structured tree.

It has also been suggested that distance between the nodes could be a vital part of synchronization and as with time, could help patterns to resonate their signals. The grid process of matching similar counts would also help here, as the linking structure between the nodes is then more uniform. Another structure is introduced that is a base ensemble repeated as a hierarchy, where both node sets are hard-coded together. If resonance can occur as part of the firing, it may help to produce those signals recognised as consciousness. Strangely, there is some biological evidence to support the structure at least. Statistical mechanics describes pairwise correlations and [15] describes a columnar architecture, rather than a widening tree-like one. As this is an automatic process, it then asks some questions. If pattern resonance is based simply on node distance, then why would a search process find it or not find it? If the construction process is automatic and some thought processes return the patterns, why would all thought processes not return the related patterns? Why is some information more difficult to find?

## 7.1 Implementation Comparisons

The whole design is based strongly on the distributed neural architecture of the human brain and as such, the author would argue that it is one of the more cognitive designs. There are a number of other models that use brain processes already and they are much more advanced than this theory. Some of these are described in [19], which explains that each firing sequence can be vectorised and measured. Then an ensemble mass can be transformed into a vector-style of structure, with weighted sets of concepts or features, but what appears to be missing from other designs is contextual information. The research of this and related 'concept tree' papers has dealt with the problem of context. It is also noted that heterogeneity in the brain is still an unknown quantity [19] and real intelligence is currently missing, but a modular design that works the same inside of each module, could be an option. As with cellular automata, the same input would produce a different result in each module. This would suggest that a summed and weighted total for a node is insufficient and it should have some other defining characteristic, as with phenotype/genotype in genetics. The understanding is stored in the structure that is saved, which is what the signal passes through. If the brain itself is modularised, the problem can initially become one of merging signals at boundaries, rather than making an intelligent decision. The task then changes from intelligent selection to interpreting the combined output correctly, which is still an unanswerable question. The design can also map to exemplars and prototypes, which is a different set of terms to describe the problem. Possibly the 'theories' term should be placed with the symbolic neural network.

## References

[1] Ackley, D.H., Hinton, G.E. and Sejnowski, T.J. (1985). A Learning Algorithm for Boltzmann Machines, Cognitive Science, Vol. 9, pp. 147-169.

[2] Anderson, J.A., Silverstein, J.W., Ritz, S.A. and Jones, R.A. (1977) Distinctive Features, Categorical Perception, and Probability Learning: Some Applications of a Neural Model, Psychological Review, Vol. 84, No. 5.

[3] Burke, R. The Wasabi Personal Shopper: A Case-Based Recommender System. In Proceedings of the 11th National Conference on Innovative Applications of Artificial Intelligence, pages 844-849. AAAI, 1999.

[4] Duan, S., Dong, Z., Hu, X., Wang, L. and Li, H. (2016). Small-world Hopfield neural networks with weight salience priority and memristor synapses for digit recognition, Neural Computing and Applications, Vol. 27, Vol. 4, pp. 837 - 844.

[5] Entree Chicago Recommendation Data Data Set, http://archive.ics.uci.edu/ml/datasets/ Entrée+Chicago+Recommendation+Data. (last accessed 14/2/17)

[6] Greer, K. (2018). New Ideas for Brain Modelling 4, BRAIN. Broad Research in Artificial Intelligence and Neuroscience, Vol. 9, No. 2, pp. 155-167. ISSN 2067-3957. Also available on arXiv at https://arxiv.org/abs/1708.04806.

[7] Greer, K. (2017). A Brain-like Cognitive Process with Shared Methods, Int. J. Advanced Intelligence Paradigms (IJAIP), Inderscience, accepted for publication, available on arXiv at http://arxiv.org/abs/1507.04928.

[8] Greer, K. (2014). Concept Trees: Building Dynamic Concepts from Semi-Structured Data using Nature-Inspired Methods, in: Q. Zhu, A.T Azar (eds.), Complex system modelling and control through intelligent soft computations, Studies in Fuzziness and Soft Computing, Springer-Verlag, Germany, Vol. 319, pp. 221 – 252, 2014. Published on arVix at http://arxiv.org/abs/1403.3515.

[9] Greer, K. (2014). New Ideas for Brain Modelling 2, in: K. Arai et al. (eds.), Intelligent Systems in Science and Information 2014, *Studies in Computational Intelligence*, Vol. 591, pp. 23 – 39, Springer International Publishing Switzerland, 2015, DOI 10.1007/978-3-319-14654-6_2.

Published on arXiv at http://arxiv.org/abs/1408.5490. Extended version of the SAI'14 paper, Arguments for Nested Patterns in Neural Ensembles.

[10] Greer, K. (2016). New Ideas for Brain Modelling, IOSR Journal of Engineering (IOSRJEN), Vol. 6, Issue 10, October, pp. 33 - 51, ISSN (e): 2250-3021, ISSN (p): 2278-8719. Also available on arXiv at http://arxiv.org/abs/1403.1080.

[11] Greer, K. (2012). Turing: Then, Now and Still Key, in: X-S. Yang (eds.), Artificial Intelligence, Evolutionary Computation and Metaheuristics (AIECM) - Turing 2012, Studies in Computational Intelligence, 2013, Vol. 427/2013, pp. 43-62, DOI: 10.1007/978-3-642-29694-9_3, Springer-Verlag Berlin Heidelberg. Published on arXiv at http://arxiv.org/abs/1403.2541.

[12] Greer, K. (2011). Symbolic Neural Networks for Clustering Higher-Level Concepts, *NAUN International Journal of Computers*, Issue 3, Vol. 5, pp. 378 – 386, extended version of the WSEAS/EUROPMENT International Conference on Computers and Computing (ICCC'11).

[13] Greer, K. (2011). Clustering Concept Chains from Ordered Data without Path Descriptions, Distributed Computing Systems, available on arXiv at http://arxiv.org/abs/1403.0764.

[14] Hämäläinen, W. and Nykänen, M. (2008). Efficient discovery of statistically significant association rules. Proceedings of the 8th IEEE International Conference on Data Mining (ICDM 2008), pp. 203-212. IEEE Computer Society.

[15] Hawkins, J. and Blakeslee, S. On Intelligence. Times Books, 2004.

[16] Hebb, D.O. (1949). The Organisation of Behaviour.

[17] Hill, S.L., Wang, Y., Riachi, I., Schürmann, F. and Markram, H. (2012). Statistical connectivity provides a sufficient foundation for specific functional connectivity in neocortical neural microcircuits, Proceedings of the National Academy of Sciences.

[18] Hopfield, J.J. (1982). Neural networks and physical systems with emergent collective computational abilities, Proceedings of the National Academy of Sciences of the USA, vol. 79, No. 8, pp. 2554 – 2558.

[19] Lieto, A., Lebiere, C. and Oltramari, A. (2017). The knowledge level in cognitive architectures: Current limitations and possible developments, Cognitive Systems Research.

[20] Muller, H.J., Elliot, M.A., Herrmann, C.S and Mecklinger, A. (2001). Neural Binding of Space and Time: An Introduction, Visual Cognition, Vol. 8, pp. 273 - 285.

[21] Murre, J.M.J and D.P.F. Sturdy, D.P.F. (1995). The connectivity of the brain: multi-level quantitative analysis, Biological cybernetics, Vol. 73, No. 6, pp. 529 - 545.

[22] Rojas, R. Neural Networks: A Systematic Introduction, Springer-Verlag, Berlin and online at books.google.com, 1996.

[23] Rosin, D.P., Rontani, D., Gauthier, D.J. and Scholl, E. (2013). Control of synchronization patterns in neural-like Boolean networks, arXiv preprint repository, http://arxiv.org.

[24] Schneidman, E., Berry, M.J., Segev, R. and Bialek, W. (2006). Weak pairwise correlations imply strongly correlated network states in a neural population, Nature Vol. 440(7087), pp. 1007 - 1012.

[25] Singer, W. and Gray, C.M. (1995). Visual feature integration and the temporal correlation hypothesis, Annu Rev Neurosci. Vol. 18, pp. 555 - 586.

[26] Tkačik, G., Mora, T., Marre, O., Amodei, D., Palmer, S.E., Berry, M.J. and Bialek, W. (2015). Thermodynamics and signatures of criticality in a network of neurons, Proceedings of the National Academy of Sciences, Vol. 112, No. 37, pp. 11508 - 11513.

[27] USDA, NRCS. 2008. The PLANTS Database (https://plants.usda.gov/java/, 31 December 2008). National Plant Data Center, Baton Rouge, LA 70874-4490 USA. http://archive.ics.uci.edu/ml/datasets/Plants (last accessed June 2017).

[28] Vogels, T.P., Kanaka Rajan, K. and Abbott, L.F. (2005). Neural Network Dynamics, Annu. Rev. Neurosci., Vol. 28, pp. 357 - 376.

[29] Watts DJ and Strogatz SH. (1998). Collective dynamics of 'small world' networks, Nature, Vol. 393, pp. 440 – 442.

[30] Weisbuch, G. (1999). The Complex Adaptive Systems Approach to Biology, Evolution and Cognition, Vol. 5, No. 1, pp. 1 - 11.

# Appendix A

The results of clustering the plants dataset [27] are listed here. The clusters are listed first, followed by the abbreviation codes for each USA or Canada state or province.

**Clusters**

1. fl, hi, pr.
2. nc, va.
3. il, in, ia, mo.
4. ky, tn.
5. la, tx.
6. md, de.
7. mi, wi.
8. ak, yt.
9. az, nm.
10. ca, nv, or.
11. co, ut, wy.
12. ga, al.
13. id, mt, wa.
14. ny, pa, ri, vt, ns, on, nj.
15. oh, wv, qc.
16. ab, bc, sk.
17. mb, nb, nt.
18. nf, nu, pe, fraspm.
19. ks, ne.
20. nd, sd, dengl.
21. ok, ar, gl.
22. ct.
23. dc.
24. ms.
25. sc.
26. vi.
27. me.
28. ma.
29. mn.
30. nh.
31. lb.

**States**

**U.S. States:**
ab Alabama
ak Alaska
ar Arkansas
az Arizona
ca California
co Colorado
ct Connecticut
de Delaware
dc District of Columbia
fl Florida
ga Georgia
hi Hawaii
id Idaho
il Illinois
in Indiana
ia Iowa
ks Kansas
ky Kentucky
la Louisiana
me Maine
md Maryland
ma Massachusetts
mi Michigan
mn Minnesota
ms Mississippi
mo Missouri
mt Montana
ne Nebraska
nv Nevada
nh New Hampshire
nj New Jersey
nm New Mexico
ny New York
nc North Carolina
nd North Dakota
oh Ohio
ok Oklahoma
or Oregon
pa Pennsylvania
pr Puerto Rico
ri Rhode Island
sc South Carolina
sd South Dakota
tn Tennessee
tx Texas
ut Utah
vt Vermont
va Virginia
vi Virgin Islands
wa Washington
wv West Virginia
wi Wisconsin
wy Wyoming

**Canada:**
al Alberta
bc British Columbia
mb Manitoba
nb New Brunswick
lb Labrador
nf Newfoundland
nt Northwest Territories
ns Nova Scotia
nu Nunavut
on Ontario
pe Prince Edward Island
qc Québec
sk Saskatchewan
yt Yukon

dengl Greenland (Denmark)
fraspm St. Pierre and Miquelon (France)